%% file: main.tex
\documentclass[sigconf]{acmart}
\AtBeginDocument{%
  }

\usepackage{multirow}
\usepackage{makecell}
\usepackage{booktabs}
\usepackage{array} 
\usepackage{graphicx}
\usepackage{rotating}
\usepackage{listings}
\usepackage{xcolor}
\usepackage{afterpage}
\usepackage{placeins}

\usepackage[caption=false,font=footnotesize,labelfont=rm,textfont=rm]{subfig}
\DeclareSubrefFormat{parens}{#1(#2)}
\setcopyright{acmlicensed}
\copyrightyear{2026}
\acmYear{2026}
\setcopyright{cc}
\setcctype{by}
\acmConference[WSDM '26] {Proceedings of the Nineteenth ACM International Conference on Web Search and Data Mining}{February 22--26, 2026}{Boise, ID, USA.}
\acmBooktitle{Proceedings of the Nineteenth ACM International Conference on Web Search and Data Mining (WSDM '26), February 22--26, 2026, Boise, ID, USA}
\acmISBN{979-8-4007-2292-9/2026/02}
\acmDOI{10.1145/3773966.3777932}
\begin{document}

\title{TableMind: An Autonomous Programmatic Agent for Tool-Augmented Table Reasoning}


\author{Chuang Jiang}
\affiliation{%
\institution{State Key Laboratory of Cognitive Intelligence, University
 of Science and Technology of China}
  \city{Hefei, Anhui Province}
  \country{China}}
\email{jiangchuang@mail.ustc.edu.cn}

\author{Mingyue Cheng}
\authornote{Mingyue Cheng is the corresponding author.}
\affiliation{%
  \institution{State Key Laboratory of Cognitive Intelligence, University of Science and Technology of China}
  \city{Hefei, Anhui Province}
  \country{China}}
\email{mycheng@ustc.edu.cn}

\author{Xiaoyu Tao}
\affiliation{%
\institution{State Key Laboratory of Cognitive Intelligence, University
 of Science and Technology of China}
  \city{Hefei, Anhui Province}
  \country{China}}
\email{txytiny@mail.ustc.edu.cn}

\author{Qingyang Mao}
\affiliation{%
\institution{State Key Laboratory of Cognitive Intelligence, University
 of Science and Technology of China}
  \city{Hefei, Anhui Province}
  \country{China}}
\email{maoqy0503@mail.ustc.edu.cn}

\author{Jie Ouyang}
\affiliation{%
\institution{State Key Laboratory of Cognitive Intelligence, University
 of Science and Technology of China}
  \city{Hefei, Anhui Province}
  \country{China}}
\email{ouyang_jie@mail.ustc.edu.cn}

\author{Qi Liu}
\affiliation{%
\institution{State Key Laboratory of Cognitive Intelligence, University
 of Science and Technology of China}
  \city{Hefei, Anhui Province}
  \country{China}}
\email{qiliuql@ustc.edu.cn}

\renewcommand{\shortauthors}{Chuang Jiang et al.}

\begin{abstract}
  \input{0-Abstract.tex}

\end{abstract}





\maketitle
\input{1-Introduction}
\input{2-RelatedWork}
\input{3-Preliminaries}
\input{4-SystemDesign}
\input{5-Experiments}
\input{6-Conclusion}

\clearpage
\section*{Ethical Considerations}
This research is based entirely on publicly accessible and fully anonymized datasets. The data utilized contains no personally identifiable information or sensitive personal details. Our study does not involve human subjects, protected attributes, or any information that could potentially infringe upon individual privacy. We have assessed the potential outcomes and foresee no adverse societal impacts stemming from this work. All procedures and methodologies employed herein strictly adhere to established ethical guidelines for responsible data handling and academic research.
\bibliographystyle{ACM-Reference-Format}
\bibliography{main}

\end{document}

%% file: 0-Abstract.tex
Table reasoning requires models to jointly perform comprehensive semantic understanding and precise numerical operations. Although recent large language model (LLM)-based methods have achieved promising results, most of them still rely on a single-turn reasoning paradigm that processes flattened tables in a single forward pass. This paradigm suffers from inherent limitations, including context overflow on large tables, weak sensitivity to continuous numerical values, and the absence of explicit tool-use and reflection. In this paper, we propose TableMind, a tuning-based autonomous programmatic table agent that simulates the human-like cognitive schema of the multi-turn interaction within a lightweight LLM. Instead of adopting a training-free workflow design, TableMind learns to internalize planning, action, and reflection through a principled two-stage training strategy. To bootstrap structured table reasoning capabilities, we construct and filter high-quality reasoning data for the supervised fine-tuning (SFT) stage. To enable precise code generation, we introduce a designed multi-perspective reward scheme and a novel optimization objective in the reinforcement learning (RL) stage. Extensive experiments on diverse benchmarks demonstrate that TableMind consistently outperforms previous baselines, validating the effectiveness of training autonomous agents to improve overall performance.  \footnote{Our codes are available at https://github.com/lennendd/TableMind.}

%% file: 1-Introduction.tex
\FloatBarrier
\begin{figure*}[htbp]
    \centering
    \includegraphics[width=1\linewidth]{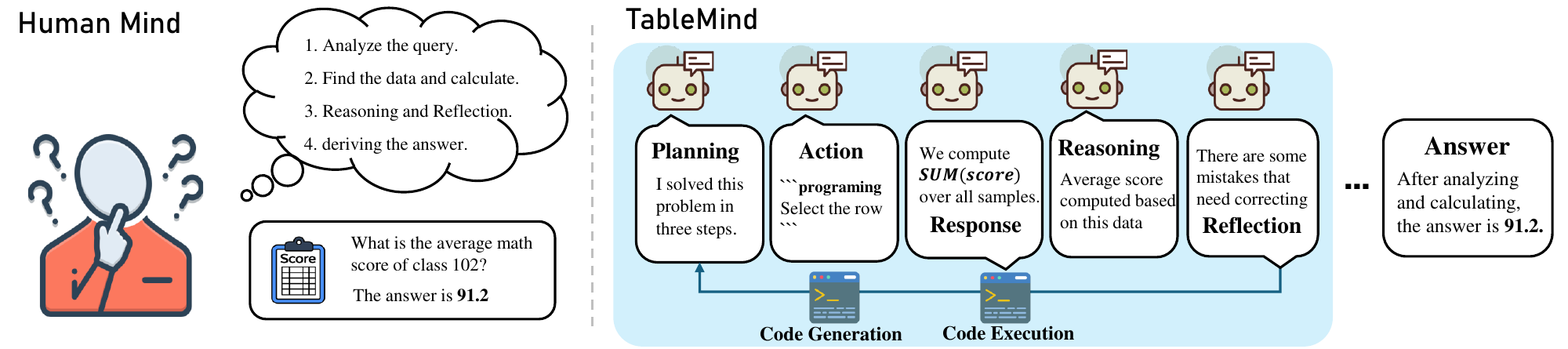}
    \caption{TableMind mimics the human chain of thought by using a multi-turn plan-action-reflect loop to solve table tasks.}
    \label{fig:overView}
\end{figure*}

\section{Introduction}
\label{sec:intro}
Tables are one of the most fundamental forms of structured data, widely used across scientific research ~\cite{hollmann2025accurate}, healthcare management~\cite{pampari2018emrqa}, and business intelligence~\cite{chen2022convfinqa}. A large amount of critical knowledge is organized in tabular formats, making table understanding and reasoning a practically important research problem. Unlike unstructured text, table reasoning requires models to jointly handle comprehensive semantic understanding and precise numerical operations.

Early studies on table reasoning primarily rely on deep learning models, which focus on task-specific representation learning for structured tables~\cite{herzig2020tapas, hollmanntabpfn}. These approaches model table structures and numerical relationships through masked pre-training or end-to-end learning paradigms.
While these methods achieve strong performance on specific tasks, they often depend on extensive labeled supervision and generalize poorly to unseen table structures. The recent advances in large language models (LLMs)~\cite{cheng2024towards, zhai2024large} introduce new opportunities for table reasoning. By leveraging rich world knowledge and strong semantic reasoning capabilities, LLM-based approaches mainly reformulate table reasoning as a language-driven generative process, enabling unified reasoning and zero-shot generalization ~\cite{zhang2024tablellm}. 

Despite their effectiveness, most existing methods still adopt a single-turn reasoning paradigm, processing tables in a single forward pass, which leads to several fundamental limitations~\cite{cheng2025survey}.  First, many LLM-based approaches~\cite{tao2025values} flatten the entire table into textual input and directly feed it to the language model. This design is vulnerable to context overflow for large tables, leading the model to reason over incomplete information, particularly in wide or long tables. Second, LLM-based methods~\cite{zhou2025benchmarking} are inherently insensitive to continuous numerical values. While they excel at semantic reasoning, they often struggle with precise numerical operations such as arithmetic, comparison, and aggregation. Treating values as plain tokens hinders reliable numerical reasoning, leading to unstable outputs and critical calculation errors. Third, existing LLM-based methods~\cite{cheng2025agent} typically lack explicit tool-use and reflection. Most existing LLM-based methods mainly depend on the intrinsic capabilities of LLMs and lack explicit mechanisms for code generation, execution, and reflection. Overall, these limitations reveal a mismatch between single-turn LLM reasoning and the multi-turn interaction process required for reliable table reasoning.

These analyses motivate a closer examination of how human cognitively perform table reasoning, as illustrated in the left part of Figure~\ref{fig:overView}. Given a query, humans typically adopt a systematic multi-stage process to derive answers from complex tables. Specifically, they analyze the query and decompose it into sub-tasks, locating relevant rows and columns, performing explicit numerical operations, checking intermediate results, and finally deriving a validated answer. A key characteristic of this process is that reasoning proceeds via explicit actions, such as calculations or numerical operations, while intermediate results are repeatedly verified~\cite{chen2019tabfact}. This observation suggests that reliable table reasoning requires explicit planning, action, and reflection, rather than a single-turn generation over structured tables.

Motivated by the human-like cognitive schema of the multi-turn interaction, a direction is to leverage LLM-driven agents~\cite{huang2024understanding} to explicitly model such processes, as shown in the right part of Figure~\ref{fig:overView}. In practice, LLMs are coordinated through predefined pipelines to perform such processes, with the most straightforward implementation being training-free workflow-based agents~\cite{yao2023react, wang2024survey}. While effective for numerical reasoning, it comes with notable limitations~\cite{cheng2024towards, zhai2024large}. Workflow-based agents typically rely on multiple LLM calls and external execution environments, resulting in computational overhead. Moreover, the need to expose tabular data to external models raises practical concerns regarding data privacy and security, especially in sensitive real-world applications. Given these limitations, we aim for a more meaningful and challenging goal: instead of relying on heavy workflows, we seek to train a lightweight table agent that intrinsically acquires human-like reasoning capabilities. Such an agent is expected to internalize planning, action, and reflection capabilities within an LLM, enabling efficient, privacy-preserving, and accurate table reasoning.

To achieve the above goal, we propose TableMind, a lightweight autonomous agent trained to internalize human-like cognitive
schema of the multi-turn interaction. TableMind is trained using a principled two-stage strategy. In the first stage, we perform supervised fine-tuning with carefully constructed high-quality data, allowing the model to learn the basic capability of table reasoning, including problem decomposition, code generation, and numerical operations. To ensure the effectiveness of this stage, we adopt a strict data filtering strategy to extract reasoning trajectories from a teacher model. In the second stage, we further introduce a reinforcement learning strategy that encourages trial-and-error exploration. We design multi-view rewards together with a novel optimization objective to promote accurate, verifiable, and reflective reasoning trajectories. Through this two-stage training process, TableMind is able to perform multi-turn interaction reasoning, including planning, automatic program generation, numerical reasoning, and reflection in a lightweight manner. Extensive experiments on diverse public benchmarks demonstrate that TableMind consistently outperforms previous baselines, validating the effectiveness of training a lightweight table agent for accurate table reasoning. 

Our main contributions are summarized as follows:
\begin{itemize}
    \item We propose TableMind, a lightweight autonomous agent that explicitly performs multi-turn interactive reasoning through planning, action, and reflection.
    \item We introduce a two-stage training strategy combining supervised fine-tuning and reinforcement learning to simulate human-like table reasoning capabilities.
    \item We conduct extensive experiments on multiple benchmarks, showing consistent improvements over previous methods.
\end{itemize}

%% file: 2-RelatedWork.tex
\section{Related Work}
With recent advances in LLMs and agents, this section focuses on research related to table reasoning and LLM agents.

\subsection{Table Reasoning}
By presenting data in a structured format, tables serve as a powerful means for organizing, storing, and analyzing information.
Inspired by the success of masked language modeling in BERT~\cite{koroteev2021bert}, involves masking parts of the table. For instance, TaPas~\cite{herzig2020tapas} adopts this by requiring the model to reconstruct masked cells during pre-training. Pasta~\cite{gu2022pasta} and TUTA~\cite{wang2021tuta} extend this concept by proposing to mask entire columns or segments of the table.
In contrast, TAPEX pursues a different paradigm. It pre-trains an encoder-decoder model on a large, synthetic SQL dataset. This enables the model to function as a SQL executor, thereby achieving a deeper understanding of the tabular structure. 
With the emergence of LLMs, the focus shifted toward general-purpose systems that can perform multiple table reasoning tasks without task-specific retraining. Models such as TableLLaMA~\cite{zhang2023tablellama}, TableLLM~\cite{zhang2024tablellm}, and the TableGPT series~\cite{zha2023tablegpt,su2024tablegpt2} adapt LLMs to structured data by combining language reasoning with table parsing, often achieving strong zero-shot and few-shot performance. 
Another line of work emphasizes workflow-driven methods. Chain-of-Table~\cite{wang2024chain} follows a data-flow paradigm, updating the table after each reasoning step, while PoTable~\cite{mao2024potable} adopts a plan-then-execute strategy, decoupling planning from execution through external tools. 
Recent reinforcement learning frameworks have notably enhanced the reasoning capabilities of LLMs, as seen in DeepSeek-R1~\cite{guo2025deepseek,luo2025time,wang2025can}. Extending this line, Table-R1~\cite{yang2025table} applies rule-based reinforcement learning to structured tables. Overall, language models have become central to table reasoning~\cite{wang2024tabletime}, underscoring their research significance.

\subsection{LLM-based Agent}
LLMs often struggle with complex reasoning or multi-step planning tasks when relying on purely text-based inference. To address these challenges, recent work has increasingly integrated external tool use and explicit planning strategies to augment model capabilities~\cite{wang2024survey}. For example, some methods adopt an iterative data-flow approach that continuously updates an internal state after each reasoning step, first generating a detailed plan and then invoking external tools to carry it out. However, such rigid, predefined pipelines can significantly limit flexibility, motivating the development of more dynamic agent frameworks that plan and select tools on the fly. One such approach is ReAct ~\cite{yao2023react}, which interleaves reasoning and action: the agent alternates between thinking and taking an action based on its current reasoning state, enabling continuous interaction with an external environment. Another example is CRITIC~\cite{gou2023critic}, where the agent learns via self-reflection by using external resources (e.g., search engines or knowledge bases) to validate and refine its decisions, effectively self-correcting mistakes. Building on these ideas, our work trains an LLM-based agent on a smaller-scale model to explore autonomous tool use. Despite progress in LLM agent research, the design of fully autonomous execution pipelines—such as enabling the agent to write and run code within a secure sandbox—remains largely unexplored.

%% file: 3-Preliminaries.tex
\begin{figure*}[ht]
    \centering
    \includegraphics[width=1\linewidth]{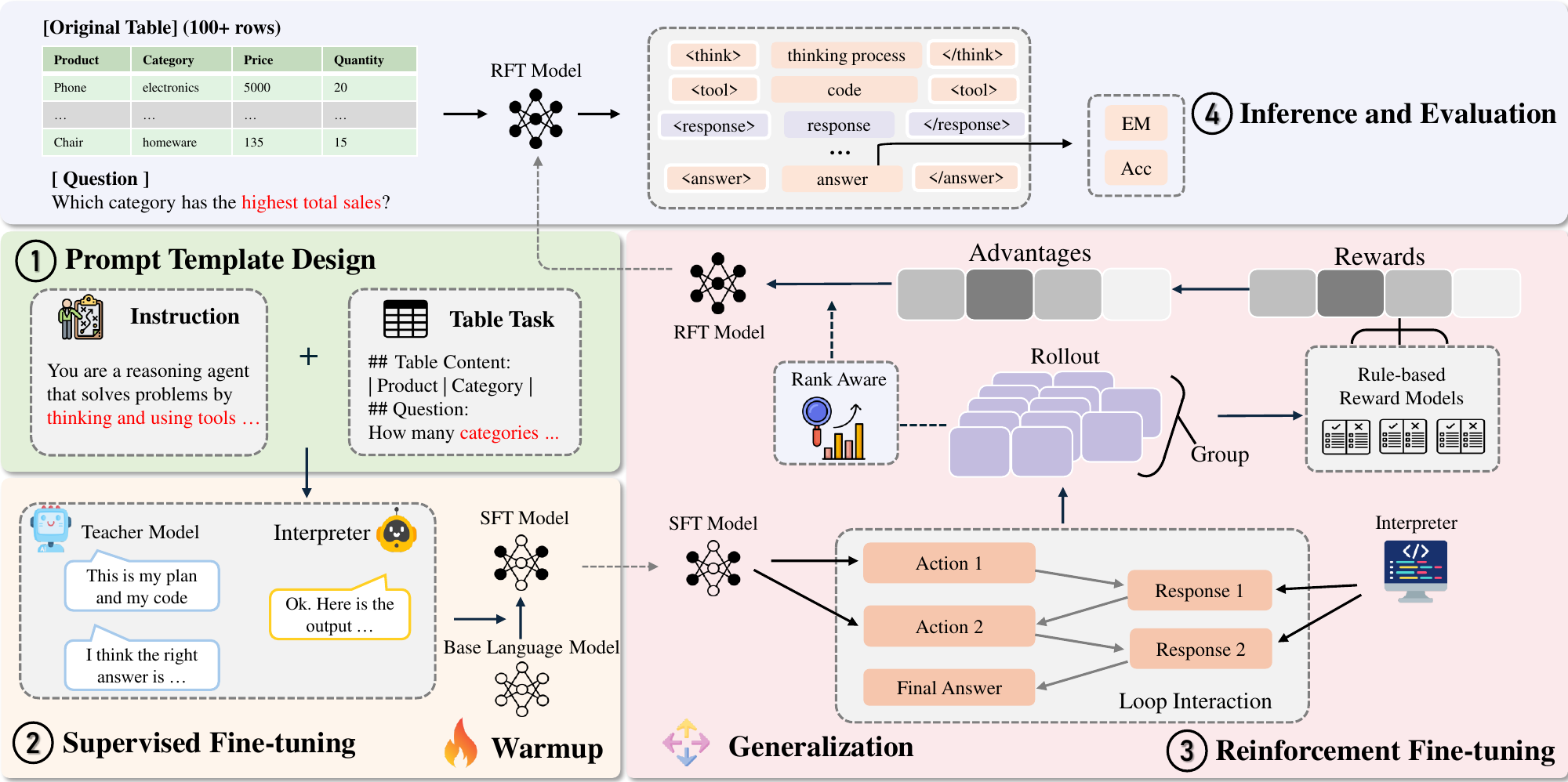}
    \caption{Overview of the multi-stage training pipeline for TableMind. The process begins with Prompt Design and SFT to warm up the model by providing it with a strong initial policy. Subsequently, RFT with RAPO is applied to significantly enhance its generalization capability. The final model is then deployed for Inference and evaluated for accuracy.}
    \label{fig:frameDiagram}
\end{figure*}
\section{Preliminaries}
In this section, we first define the table reasoning problem and formalize its key components. We then present the core ideas and underlying principles of TableMind that guide our approach.
\subsection{Task Formulation}
We consider the task of table reasoning as a structured decision-making problem grounded in tool-augmented language modeling. Each instance is represented as a tuple \((T, Q, A)\), where \(T\) denotes a structured table, \(Q\) is a natural language query (e.g., a question or a factual statement), and \(A\) is the corresponding answer. Unlike conventional QA settings, table reasoning often involves multi-step computation, intermediate result verification, and error correction. Thus, the task requires not only semantic understanding of  \(T\) and \(Q\), but also precise numerical reasoning, which may involve intermediate state tracking, external tool usage, and dynamic adjustment of the reasoning path. Formally, the goal is to learn a reasoning policy \(\pi\) that, given the initial inputs \((T, Q)\), generates a sequence of intermediate reasoning steps \(\{s_k\}_{k=1}^{K}\), and produces a final answer \(\hat{A} = \mathcal{R}(\{s_k\}_{k=1}^{K})\), where \(\mathcal{R}\) denotes an aggregation or termination function, and each step \(s_k\) may involve code generation, execution, and feedback-based refinement.

\subsection{Design Principles}
LLMs demonstrate strong capabilities in understanding natural language, making them valuable tools for the complex task of table reasoning~\cite{zhang2020summarizing, zhang2025star}. This task requires interpreting structured data, executing precise computations, and synthesizing intermediate results, often necessitating interaction with external tools to ensure accuracy. However, current research presents a distinct dilemma, forcing a choice between two insufficient paradigms.
These methods lack mechanisms for external verification or self-correction, often failing on tasks that require complex computation. On the other hand, the few methods that do incorporate tool interaction~\cite{masterman2024landscape} are heavily reliant on massive, proprietary models. Their tool invocation patterns are often rigid and pre-defined, lacking the capacity for dynamic reflection, and their dependence on large-scale models creates significant barriers to computational cost and adaptability.
This landscape reveals a critical and unfilled research gap: the absence of systems that combine autonomous tool invocation and reflection with lightweight, accessible models. 

%% file: 4-SystemDesign.tex
\section{TableMind}
This section presents the training pipeline of TableMind. It begins with the prompt template design, followed by the two-stage training strategy, and concludes with inference and evaluation.
\subsection{Overview}
An overview of TableMind is shown in Figure~\ref{fig:frameDiagram}. We first design a Prompt Template to standardize diverse table inputs and task descriptions, guiding the model to reason by combining thinking and tool usage.
TableMind uses a two-stage training strategy. The supervised fine-tuning (SFT) stage equips the model with basic tool-invocation and reasoning capabilities using trajectories distilled from teacher model. The reinforcement fine-tuning (RFT) stage allows the model to autonomously explore and refine reasoning strategies. To improve stability in reasoning tasks, we introduce a rank-aware policy optimization (RAPO) algorithm and a multi-perspective reward function, aligning model confidence with trajectory quality for stable policy improvement.
During Inference and Evaluation, the model executes iterative reasoning over tables, generating actions, and deriving final answers. 

\subsection{Prompt Template Design}
We propose a structured prompt template to ensure consistent and reliable reasoning for table-based tasks, aiming to reduce ambiguity in table understanding and to provide a stable interface for reasoning and tool usage.
The prompt integrates high-level reasoning guidance with task-specific information to address the inherent complexity of table reasoning, which requires coordinated data interpretation, problem decomposition, and tool invocation. Concretely, it consists of a general Instruction block that defines the model’s role and enforces a fixed reasoning workflow, and a Table Task block that provides the table content and corresponding query. This design yields more stable reasoning trajectories and more accurate tool usage across diverse table reasoning tasks.

\subsection{Supervised Fine-tuning}
The SFT stage primarily warms up the agent’s interaction behavior by distilling high-quality trajectories and aligning the model with a structured reasoning paradigm.
\subsubsection{Data Collection}
We construct a high-quality dataset of reasoning trajectories for SFT using an iterative, multi-turn framework. At its core, this framework employs a knowledge distillation~\cite{hinton2015distilling} approach to transfer the programmatic knowledge and reasoning processes of an expert model into structured training samples. Specifically, we prompt the expert model to generate a plan and an initial executable action, which is then executed in a local environment. The resulting output or error message is returned to the model as an observation, enabling it to reflect on the outcome, refine its plan, and generate the next action. This interactive loop continues until the task is completed and a final answer is derived. All collected trajectories undergo a rigorous calibration process: we compare the final output of each trajectory against a predefined ground-truth answer and retain only those that lead to correct solutions. This filtering ensures that the resulting dataset consists exclusively of high-quality, validated reasoning paths.

\subsubsection{Warm-up Training Stage}
The SFT stage serves as a warm-up phase that aligns the model’s output format, enables it to master the basic syntax of the planning--action--reflection loop, and establishes an initial policy for generating valid and logically coherent code. This initial training phase acts as a preparatory step, substantially lowering the optimization overhead for subsequent RFT. However, while SFT successfully imparts foundational instruction-following capabilities and reasoning skills, the resulting reasoning behavior is not yet consistent. It primarily reflects pattern memorization from the training data and exhibits limited generalization to out-of-distribution tasks. Consequently, the reasoning ability induced by SFT can be viewed as a form of non-generalizable memory, necessitating further refinement through RFT.

\subsection{Reinforcement Fine-tuning}
Building upon the warm-up provided by SFT, the RFT stage improves generalization by optimizing the agent with carefully designed reward functions and a policy optimization objective.
\subsubsection{Multi-perspective Reward Design}
We carefully design a multi-perspective reward function, composed of three key components to effectively guide the TableMind agent in exploring tool-use strategies and enhancing the efficiency of RLVR~\cite{yue2025does}. 
To ensure that the agent produces syntactically valid and reasoning trajectories, we define a format reward $R_{\mathrm{format}}$ to enforce structural integrity. 
\begin{equation}
R_{\mathrm{format}} =
\begin{cases}
1, & \text{if output format is valid}, \\
0, & \text{otherwise}.
\end{cases}
\label{eq:format_reward}
\end{equation}

A binary reward is assigned based on whether the output adheres to the required structured format.
 As the primary indicator of task success, the accuracy reward $R_{\mathrm{acc}}$ 
directly assesses the correctness of the agent's final answer. The reward is defined as:

\begin{equation}
R_{\mathrm{acc}} =
\begin{cases}
1, & \text{answer matches the ground truth}, \\
0, & \text{otherwise}.
\end{cases}
\label{eq:accuracy_reward}
\end{equation}

For table question, correctness is measured by exact match with ground-truth answers, while for table fact verification, it is measured by label accuracy. To guide efficient and effective tool use, we introduce an auxiliary reward $R_{\mathrm{tool}}$ with an implicit curriculum that encourages exploration early and efficiency later.
\begin{equation}
R_{\mathrm{tool}} = e^{-\rho s} \left( \beta \cdot I_{\mathrm{success}} 
- C \cdot (N_{\mathrm{turns}})^{2} \right),
\label{eq:tool_reward}
\end{equation}
where $s$ is the number of global training steps, $\mathbb{I}(\cdot)$ is the indicator~function that returns $1$ if at least one tool call in the trajectory is successful, $\beta$ is the positive base reward, $N_{\mathrm{turns}}$ is the number of tool turns, and $C$ is a penalty coefficient. The total reward for each reasoning trajectory is the sum of the above components.

\subsubsection{RAPO}
We introduce RAPO, a method to improve group-based policy gradient algorithms. It identifies misaligned trajectories and boosts their learning signal using rank-aware advantage weighting.
The RAPO objective modifies the original GRPO~\cite{shao2024deepseekmath} framework and incorporates some frontier techniques. First, following recent work~\cite{hu2025open, liu2025understanding}, we removes the KL divergence, allowing the model to no longer be constrained by the reference model and thereby expanding the search space. Second, it employs a token-level policy gradient loss to reduce the impact of answer length on the gradient. Finally, it uses the clip-higher strategy, which promotes system diversity and prevents entropy collapse. The objective function is formalized as:
\begin{equation}
\begin{aligned}
J_{\mathrm{RAPO}}(\theta) 
&= \mathbb{E}_{q, \{o_i\}_{i=1}^G \sim \pi_{\theta_{\mathrm{old}}}} \bigg[\quad \frac{1}{\sum_{i=1}^G |o_i|} \sum_{i=1}^G 
\sum_{t=1}^{|o_i|} 
\min\big( r_{i,t}(\theta) A'_i,\; \\
&\mathrm{clip}(r_{i,t}(\theta), 1-\epsilon_{\mathrm{low}}, 1+\epsilon_{\mathrm{high}}) A'_i \big)
\bigg].
\end{aligned}
\end{equation}

Here, $r_{i,t}(\theta)$ is the probability ratio between the current policy $\pi_{\theta}$ and the old policy $\pi_{\theta_{\text{old}}}$ that was used for data collection:
\begin{equation}
\begin{aligned}
r_{i,t}(\theta) = \frac{\pi_{\theta}(o_{i,t} \mid q, o_{i,<t})}{\pi_{\theta_{\mathrm{old}}}(o_{i,t} \mid q, o_{i,<t})}.
\end{aligned}
\end{equation}

Building upon this framework, we replace the standard group-normalized advantage with the proposed dynamically weighted, rank-aware advantage $A'_i$. This change ensures that the gradient signal for each token is not only directly proportional to the trajectory’s quality but also explicitly proportional to the degree of misalignment between the model’s confidence and that quality.
\begin{equation}
\begin{aligned}
A'_i = \gamma_i \cdot \frac{R_i - \mathrm{mean}(\{R_j\}_{j=1}^G)}{\mathrm{std}(\{R_j\}_{j=1}^G)}.
\end{aligned}
\end{equation}

The central mechanism of RAPO is its method for identifying the learning signal from misaligned trajectory pairs. A misalignment occurs when the model assigns a higher confidence to a low-reward trajectory $o_l$ than to a high-reward trajectory $o_w$ from the same group. This confidence is measured by the log-probability of the sequence, $\log P(o_i)$, which is the length-normalized log-probability of its tokens.
To detect these misalignments, we introduce a pairwise weighting factor $\gamma_{w,l}$, which acts as a diagnostic flag for each winner loser pair. If the winner's log-probability does not exceed that of the loser, the pair is considered misaligned, and this factor is elevated to apply extra learning pressure. It is defined as:
\begin{equation}
\gamma_{w,l} = 1 + \alpha \cdot \mathbb{I}[\log P(o_w) < \log P(o_l)],
\end{equation}
where $\mathbb{I}[\cdot]$ is the indicator function and $\alpha$ is a hyperparameter controlling the intensity of the re-weighting.

\subsection{Inference and Evaluation}
During the inference phase, the model employs an iterative reasoning framework to solve complex problems. This framework operates in a multi-turn loop, where each iteration consists of several core components. The model first analyzes the user query together with the current state to formulate a strategy for the next action. It then generates executable Python code to query or process the tabular data, which is securely executed by a code interpreter to produce an output. The execution result is subsequently provided to the model as an observation, based on which the model reflects to assess progress, verify correctness, and determine whether to generate the final answer or continue to the next iteration.This planning–action–reflection loop enables the model to perform multi-turn reasoning, leverage intermediate results, and self-correct when necessary. The process terminates either when the model determines during the reflection step that a solution has been reached or when a predefined maximum number of iterations is exceeded. Finally, the model synthesizes all information accumulated throughout the iterations to produce the final answer.
For evaluation, we adopt exact match as the metric for table question answering tasks and accuracy for table fact verification tasks.

%% file: 5-Experiments.tex
\definecolor{color1}{HTML}{DDEBF7} 
\definecolor{color2}{HTML}{BDD7EE} 
\definecolor{color3}{HTML}{5B9BD5} 
\definecolor{color4}{HTML}{2F75B6} 
\definecolor{color5}{HTML}{204E79} 

\newcommand{\legenditem}[2]{%
    \colorbox{#1}{\rule{0pt}{2pt}\rule{3pt}{0pt}}\hspace{0.5em}#2%
}

\begin{table}[h]
    \centering
    \caption{Dataset statistics for training and evaluation. In-domain datasets (TabFact, TabMWP, WTQ) are used for both training and evaluation, while out-of-domain datasets (HiTab, FinQA) are used only for evaluation.}
    \label{tab:dataset_stats}
    \begin{tabular}{lccccc}
        \toprule
        \textbf{Split} & \textbf{TabFact} & \textbf{TabMWP} & \textbf{WTQ} & \textbf{HiTab} & \textbf{FinQA} \\
        \midrule
        Training   & 3,500 & 3,500 & 1,000 & -- & -- \\
        Test (RL)  & 800  & 800 & 800 & -- & -- \\
        Evaluation & 12,779  & 7,686 & 4,344 & 1,349 & 1,138 \\
        \bottomrule
    \end{tabular}
\end{table}

\input{table/1-main-result}

\section{Experiments}
In this section, we outline the experimental setup of TableMind, including datasets, baselines, and evaluation metrics, followed by an analysis of the results to evaluate its effectiveness.
\subsection{Experimental Settings}
\subsubsection{Datasets.} 
Table~\ref{tab:dataset_stats} presents detailed dataset statistics. We employ three in-domain datasets for training and evaluation: WikiTQ, HiTab, and FinQA, constructing high-quality slow-thinking demonstrations using the answer-aware generation approach described in Section~3.2.
We evaluate TableMind on comprehensive table reasoning benchmarks, categorized into in-domain and out-of-domain evaluations to fully assess the model's performance and generalization capabilities.
For in-domain evaluation, we utilize three widely used public datasets: WikiTQ~\cite{pasupat2015compositional}, TabMWP~\cite{lu2022dynamic}, and TabFact~\cite{chen2019tabfact}.
WikiTQ focuses on open-domain question answering by generating short answers from Wikipedia tables and their associated text.TabMWP targets multi-step mathematical reasoning over tables, requiring accurate numerical computation.TabFact addresses fact verification by judging whether a statement is supported by a given Wikipedia table.
To further evaluate the robustness and transferability of TableMind, we extend our experiments to two out-of-domain datasets: HiTab~\cite{cheng2021hitab} and FinQA~\cite{chen2021finqa}.
HiTab introduces hierarchical tables that require understanding complex header dependencies for statistical question answering.FinQA targets expert-level financial reasoning, requiring multi-step numerical calculations over both tables and text.
All evaluations are conducted on the test sets of the respective datasets to ensure consistency and fair comparison with prior work.

\subsubsection{Baselines.} 
We choose six strong LLM-based methods as baselines for evaluating three table reasoning tasks~\cite{zhao2024tapera, ye2023large}. Tab-CoT~\cite{jin2023tab} proposes a method that prompts a LLM to generate a table that models a complex reasoning process in a structured, step-by-step manner. PoTable~\cite{mao2024potable} employs a plan-then-execute two-stage framework, which is strictly followed by external tools to solve the tabular task. Chain-of-Table~\cite{wang2024chain} introduces a paradigm centered on evolving tables, where the table is updated after each reasoning step and serves as the input for the subsequent operation, thus evolving throughout the reasoning chain. TableLlama~\cite{zhang2023tablellama} fine-tunes the Llama 2 model on a diverse table-instruction dataset named TableInstruct, enabling it to handle complex table tasks without requiring a special model architecture.
TableGPT2~\cite{su2024tablegpt2} introduces a semantic table encoder to capture table structure and content, and leverages pre-training and fine-tuning for deep tabular understanding. Table-R1~\cite{yang2025table} is a reasoning model trained specifically for table tasks using reinforcement learning with verifiable rewards, to enable effective inference-time scaling.

\subsubsection{Implementation Details.}
For TableMind, we adopt Qwen3-8B as the backbone model. During the supervised fine-tuning stage, we carefully train on 200 synthetic samples with a learning rate of $1 \times 10^{-6}$ for one epoch. In the RFT stage, we explicitly implement the RAPO algorithm using the agent-r1 framework, with vLLM consistently employed for generation\footnote{https://github.com/0russwest0/Agent-R1}. In Eq.~\eqref{eq:tool_reward}, the hyperparameters are set as $\rho = 0.05$, $C = 0.01$, $\beta = 0.5$, and the group size $G = 8$. The batch size is 128, the learning rate is $1 \times 10^{-6}$, the policy temperature is 1, the maximum tool-calling turns per prompt is 3, and the maximum response length for each turn is 2048 tokens. Both training stages are extensively conducted on a 4-GPU A800 cluster.
To ensure a strictly fair comparison of generation performance, we standardize the number of samples to 1 across all baselines.

\subsection{Table Reasoning Performance} To evaluate the effectiveness of TableMind, we conduct a comprehensive comparison against competitive baselines across five diverse benchmarks, categorized into in-domain tasks (Wikitq, TabMWP, TabFact) and out-of-domain tasks (HiTab, FinQA). The quantitative results, summarized in Table~\ref{tab:main result}, demonstrate that TableMind achieves the best performance across all five datasets. Notably, it consistently surpasses strong open-source general LLMs and specialized table models across both in-domain and out-of-domain benchmarks. Furthermore, TableMind exhibits strong generalization capabilities on the unseen HiTab and FinQA datasets, highlighting its adaptability to complex structural reasoning.
Unlike Tab-CoT, which relies solely on textual reasoning, PoTable and Chain-of-Table leverage external tools to minimize calculation errors, accounting for their improvements over standard CoT. Among tuning-based methods, while TableLlama and TableGPT2 rely on SFT to imitate static reasoning paths, Table-R1 employs a dynamic RFT framework, allowing it to surpass SFT-based limitations. It is worth noting that the strong general reasoning model Deepseek-R1 also shows competitive performance, securing the runner-up position on TabMWP and HiTab. However, our approach, TableMind, unifies the strengths of tool-augmented execution and the RLVR training paradigm. This dual strategy allows TableMind to surpass Deepseek-R1 and Table-R1 on both the complex hierarchical tables of HiTab and the financial reasoning tasks of FinQA. 

\FloatBarrier
\begin{figure*}[htbp]
    \centering
    \includegraphics[width=1\linewidth]{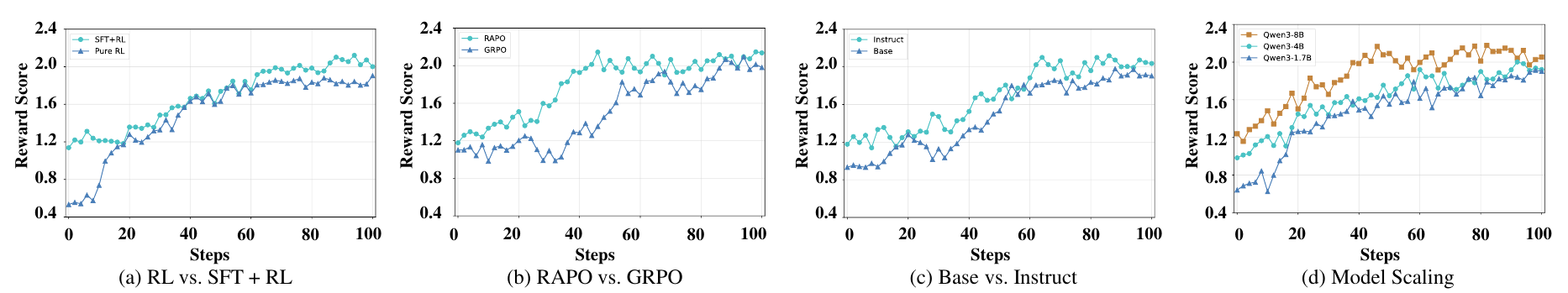}
    \caption{Analysis of Reward Curves for Key Model Components. (a) SFT initialization provides a significant advantage over Pure RL. (b) Our RAPO algorithm outperforms the GRPO baseline. (c) An instruct-tuned model learns more effectively than a base model. (d) Reward scores improve consistently with model scaling.}
    \label{fig:reward}
\end{figure*}

\input{table/2-pure-sft}
\subsection{Ablation Studies of Key Components}
\subsubsection{Effect of Supervised Fine-tuning.}
To determine the optimal strategy for the SFT stage, we conducted experiments on various training dataset sizes and epochs. The results are presented in Table~\ref{tab:pure sft}. They indicate that the optimal performance is achieved with a sample size of 200 at one epoch. While performance scales with the dataset size, increasing the training to two epochs degrades results, suggesting overfitting on our relatively small dataset. 
As illustrated by the reward curve in Figure~\ref{fig:reward} (a), employing SFT as a warm-up phase enhances the efficiency of the subsequent RFT process. Compared to a pure RFT approach, the SFT-initialized model achieves a higher initial reward score during the early stage of training. As shown in Figure ~\ref{fig:Ablation}, the removal of the SFT phase leads to a degradation in model performance. This finding substantiates the necessity of SFT for our training methodology.

\subsubsection{Effect of Reinforcement Learning.}
In this study, we remove the RFT training stage and directly evaluate the model trained only with SFT. As shown in Figure~\ref{fig:Ablation}, this modification leads to a notable performance drop across benchmarks. On both WikiTQ and TabFact, the model’s accuracy decreases substantially, with similar declines observed across other evaluation metrics. These results indicate that while SFT provides a solid starting point by establishing baseline reasoning and code-generation skills, it does not equip the model with the adaptability and decision-making depth required for complex table reasoning tasks.
The RFT phase plays a crucial role in this regard by exposing the model to execution-based feedback, enabling it to refine tool-use policies and recover from intermediate errors.  Consequently, we identify RFT as a critical technique for boosting model performance, substantially enhancing generalization and enabling the model to maintain high reliability across diverse table reasoning scenarios.

\subsubsection{Effect of the Multi-perspective Reward Design.}
We analyze the impact of our multi-objective reward function, with a particular focus on the contribution of our proposed Strategic Interaction Reward $R_{\text{tool}}$. To do so, we conducted an ablation study by removing this auxiliary component, training an agent that relies solely on the primary accuracy $R_{\text{acc}}$ and format rewards $R_{\text{format}}$. As shown in Figure~\ref{fig:Ablation}, removing the $R_{\text{tool}}$ term leads to a degradation in performance across all benchmarks.
Without the guidance of $R_{\text{tool}}$, the agent tended to generate less efficient and often erroneous reasoning chains. Specifically, the absence of the success incentive and efficiency penalty in $R_{\text{tool}}$ resulted in longer, more redundant tool-call sequences that were less likely to contribute to the correct final answer. The implicit curriculum provided by the reward's decay weight was also lost, leading to slower convergence.

\subsubsection{Effect of RAPO}
\begin{figure}[t]
    \centering
    \includegraphics[width=1\linewidth]{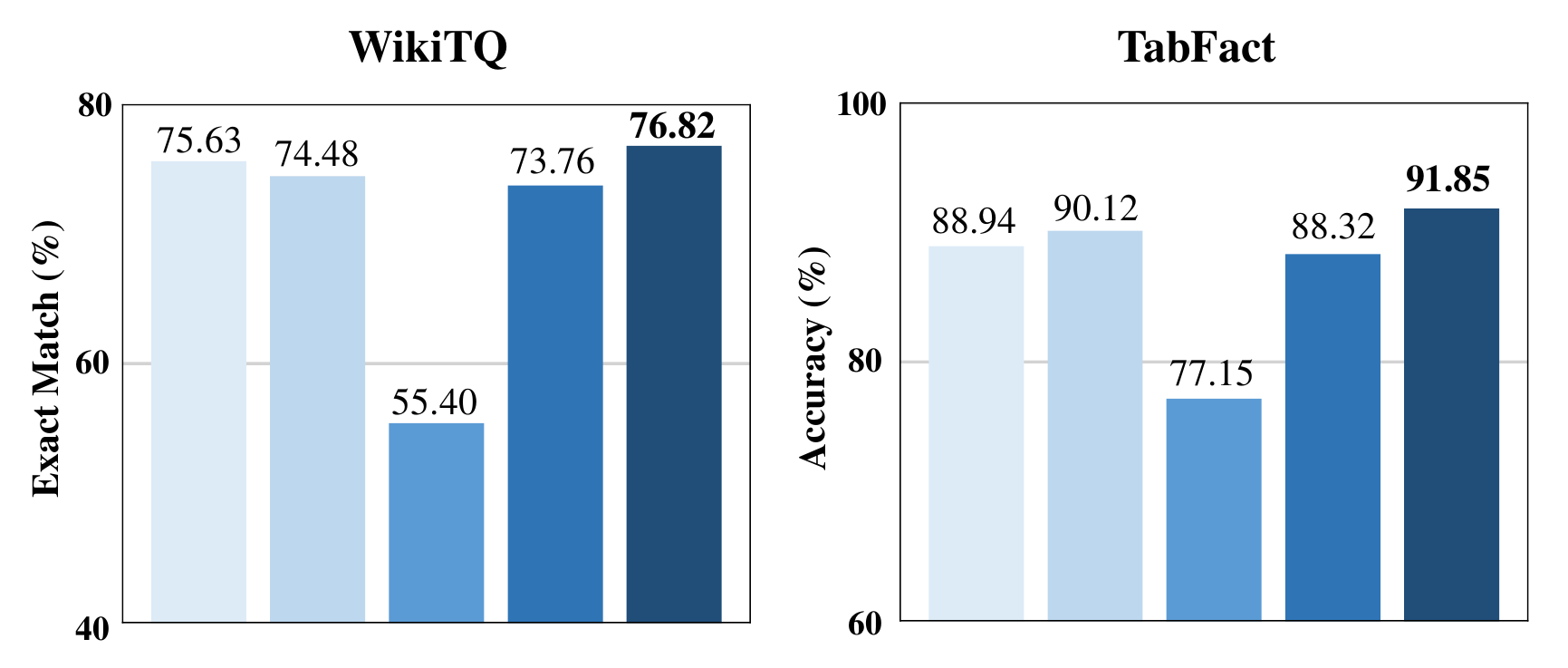}
    
    {\small
        \legenditem{color1}{w/o $R_{\text{tool}}$} \hspace{0.5em}
        \legenditem{color2}{w/o RAPO} \hspace{0.5em}
        \legenditem{color3}{w/o RFT} \hspace{0.5em}
        \legenditem{color4}{w/o SFT} \hspace{0.5em}
        \legenditem{color5}{Original}
    }
    
    \caption{Ablation study of model components. Performance on WikiTQ and TabFact after removing key components, compared to the full model.}
    \label{fig:Ablation}
\end{figure}

Ablation studies were performed from two key perspectives to thoroughly assess the effectiveness of our proposed RAPO  algorithm. First, we compared the training dynamics of RAPO against the baseline GRPO algorithm. The reward score curves in Figure~\ref{fig:reward} (b) illustrate this comparison. Throughout the training process, the reward curve for RAPO is higher than that of GRPO. This indicates that RAPO not only converges faster but also exhibits a more stable training process with less volatility, demonstrating its advantages in improving the sample efficiency of reinforcement learning. Second, we quantified RAPO's contribution to the final task performance by removing it from the full model. As shown in Figure~\ref{fig:Ablation}, removing RAPO led to a degradation in performance on both benchmark datasets. The performance decreased on both the WikiTQ and TabFact datasets, showing a noticeable drop in exact match on WikiTQ and in accuracy on TabFact.

\subsection{Comparison of Model Variants}
\subsubsection{Effect of Instruction Tuning.}
Instruction tuning is hypothesized to align a foundational model’s general capabilities with the specific demands of following task-oriented commands, thereby providing a more favorable starting point for downstream optimization. We examine this hypothesis by comparing an Instruct-tuned model against a base model within our RFT framework. The reward curves in Figure~\ref{fig:reward} (c) clearly validate this claim. The Instruct model not only starts with a substantial reward advantage—reflecting its prior adaptation to structured, goal-directed instructions—but also sustains this superiority throughout the entire training process, ultimately converging to a higher and more stable reward plateau. This behavior suggests that pre-aligning the model yields a more effective initial policy, reduces the exploration burden during RFT, and accelerates the acquisition of complex behaviors by enabling the agent to focus on fine-tuning execution strategies rather than learning fundamental task-following skills from scratch.

\subsubsection{Effect of Model Scaling.}
To understand the relationship between model capacity and performance on our task, we investigate the effect of model scaling. We apply our full training pipeline to three Qwen3 models of increasing size: 1.7B, 4B, and 8B parameters. The results, shown in Figure~\ref{fig:reward} (d), reveal a strong, positive correlation, consistent with the established scaling laws of LLMs. A clear performance hierarchy emerges from the very beginning of training and remains stable through convergence, with larger models consistently achieving higher rewards at every stage. This pattern suggests that greater representational capacity enables the model to capture more complex reasoning patterns, maintain richer intermediate state representations, and more effectively coordinate the multi-step processes involved in tool manipulation. These observations indicate that model scaling is not only a direct path to performance improvement but also a critical factor in mastering high-complexity reasoning tasks in this domain.

\begin{figure}[t]
    \centering
    \includegraphics[width=1\linewidth]{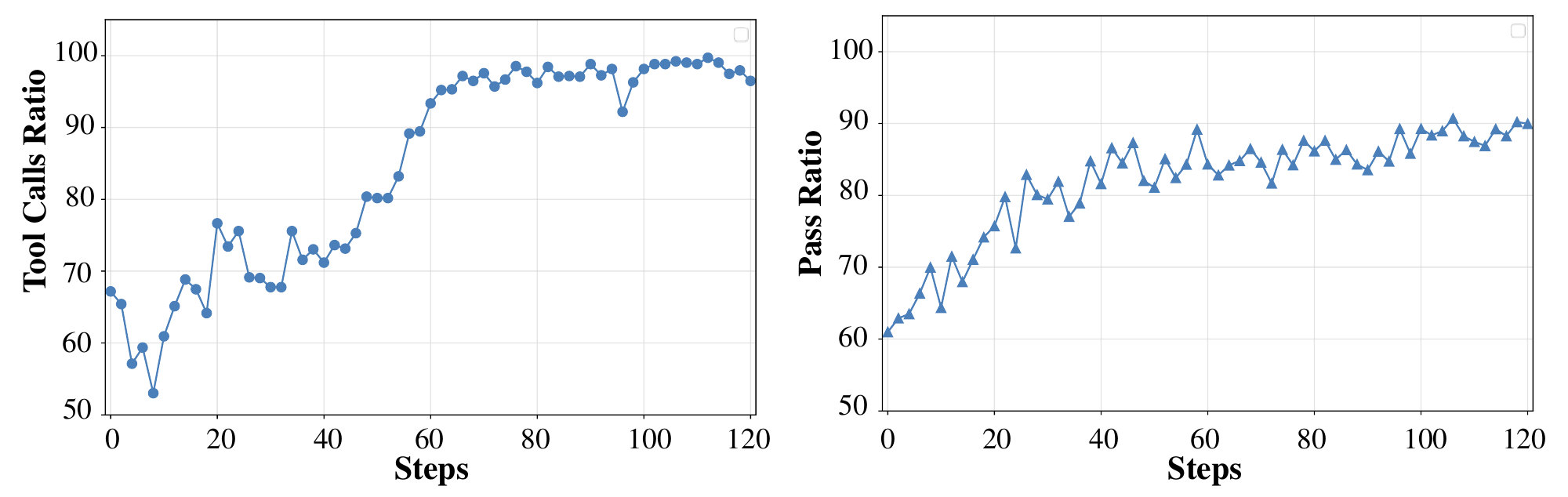}
    \caption{Evolution of the Tool Calls Ratio (left) and Pass Ratio (right) during reinforcement learning.}
    \label{fig:toolCall_lineChart}
\end{figure}
\subsection{Analysis of Tool-use}
\subsubsection{Tool Calls Ratio Analysis.}
The evolution of the agent's policy for invoking tools is shown in Figure~\ref{fig:toolCall_lineChart} (left). The learning trajectory exhibits a clear and continuous progression with distinct phases. Initially, during the first 40 steps, the agent undergoes an exploration phase, where its commitment to using tools is tentative and fluctuates as it balances direct textual generation with tool interaction, occasionally reverting to text-only outputs when tool use does not appear immediately beneficial. This is followed by a period of rapid discovery and exploitation, spanning steps 41 to 70, marked by a sharp inflection in the curve. Finally, during steps 71 to 120, the agent's strategy converges, with the tool-use ratio stabilizing near 100\%, indicating that the model has learned a  policy that treats tool use as the optimal approach for the vast majority of tasks in this domain and applies it with high confidence. 

\subsubsection{Pass Ratio Analysis.}
The pass ratio, illustrated in Figure~\ref{fig:toolCall_lineChart} (right), quantifies the reliability of the agent’s generated code, measured as the proportion of executions producing the expected results without errors. The curve shows a clear and sustained improvement in code-writing proficiency over the course of RFT training, indicating progressive refinement of the agent’s coding strategy.
The model begins with a baseline pass ratio of approximately 60\%, reflecting the syntactic and semantic code generation capabilities developed during the SFT stage. During the RFT phase, the pass ratio steadily rises to nearly 90\%, with execution-based feedback guiding the agent to reinforce correct coding patterns, eliminate common sources of runtime errors, and align code more closely with task-specific requirements. These results suggest that the combination of SFT and RFT produces an agent that not only decides when to invoke a tool but also executes the invocation reliably and consistently across different problem settings.

\input{table/3-Hyperparameter}

\FloatBarrier
\begin{figure*}[htbp]
    \centering
    \includegraphics[width=1\linewidth]{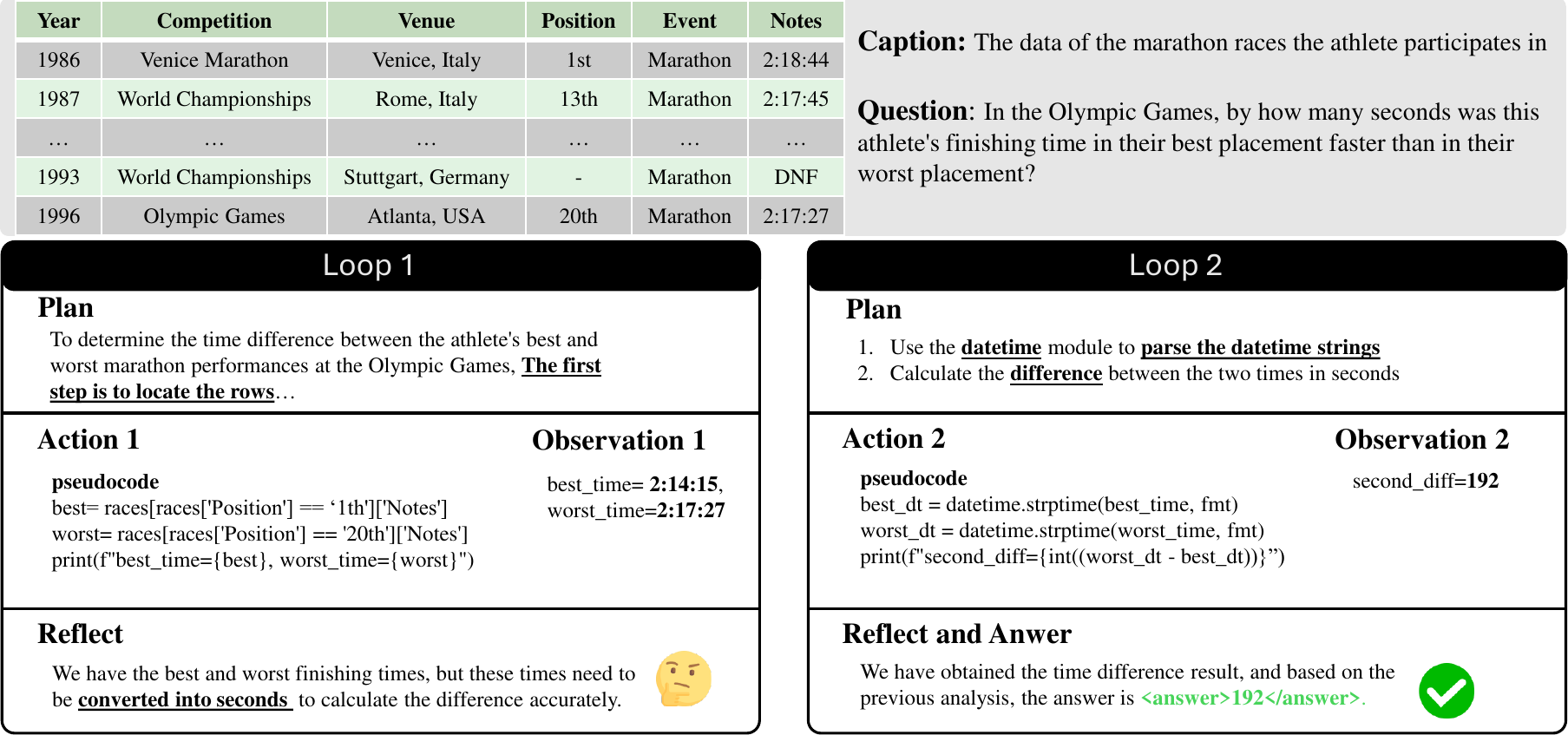}
    \caption{A case study demonstrates the model's plan-action-reflect loop. By breaking down the problem into several steps and making multi-turn tool calls, the model ultimately solves the task.}
    \label{fig:case}
\end{figure*}

\subsection{Hyperparameter Sensitivity Study}
We conduct a targeted study to identify the optimal settings for key hyperparameters and evaluate how sensitive our model is to their variations. This study focused on two core hyperparameters, the maximum number of interaction turns and the generation temperature. The results are summarized in Table~\ref{tab: hyperparam}. The max turns parameter defines the maximum number of thought-action loops the agent can perform to solve a problem. The results  indicate that: model performance significantly improves across all three datasets when max turns is increased from 1 to 3. This suggests that a single turn is insufficient for solving the complex, multi-step reasoning required by these tasks. However, when max turns is further increased from 3 to 5, a slight degradation in performance is observed. This is likely because excessive turns can increase the risk of error propagation or the generation of redundant, unhelpful steps, which can interfere with the final outcome. Based on these findings, we identify 3 as the optimal value for max turns. The temperature parameter controls the randomness of the model's generated outputs. A lower temperature leads to more deterministic responses, while a higher temperature encourages diversity. The results in the lower half of the table show that, within the tested range, model performance is positively correlated with temperature.

\subsection{Case Studies}
We present a detailed case study in Figure~\ref{fig:case} that reveals the inner workings of TableMind on a representative WikiTQ problem. The model begins by formulating a simple initial plan, successfully extracting two time strings from the table. Crucially, instead of proceeding with a naïve computation that would yield an incorrect result, the model enters a reflection state. In this stage, it identifies that a simple subtraction is infeasible because the time data is stored as textual strings rather than numerical values. Armed with this insight, the model devises a refined plan: it first converts the textual strings into a proper time format suitable for arithmetic operations, and then computes the precise difference. To implement this plan, the model generates executable code using python's datetime module. The code runs flawlessly, producing the correct numerical output of 192 seconds. Finally, the model synthesizes this raw numerical result into a well-formatted answer, ready to be returned. This example vividly illustrates that TableMind is capable of decomposing complex queries, reflecting on intermediate steps, self-correcting its reasoning strategy, and strategically leveraging external tools to solve table-based problems.

%% file: table/1-main-result.tex
\begin{table*}[htbp]
  \centering
  \caption{Main results (\%) of different baselines over the benchmarks. In each column, the best result is marked in bold and the second-best result is underlined, while the relative improvement of TableMind over the runner-up is recorded in parentheses.}
  \setlength{\tabcolsep}{12pt}

    \begin{tabular}{llccc >{\centering\arraybackslash}p{1.2cm} >{\centering\arraybackslash}p{1.2cm}}
    \toprule
    \multirow{2}{*}{} & \multirow{2}{*}{Models} & \multicolumn{3}{c}{In-domain Evaluation} & \multicolumn{2}{c}{Out-of-domain Evaluation} \\
    \cmidrule(lr){3-5} \cmidrule(lr){6-7}
          &       & Wikitq & TabMWP & TabFact & HiTab & FinQA \\
    \midrule
    \multirow{2}{*}{Open-source} & Qwen2.5-72B-Instruct & 68.56 & 97.45 & 87.34 & 68.02 & 35.83 \\
          & Deepseek-R1 & 74.63 & \underline{98.03} & 86.25 & \underline{70.37} & 37.42 \\
    \midrule
    \multirow{3}{*}{Training-free} & Tab-CoT & 60.43 & 92.21 & 79.05 & 60.64 & 27.81 \\
          & PoTable & 69.56 & 95.76 & \underline{88.93} & 61.96 & 24.93 \\
          & Chain-of-Table & 68.31 & 96.17 & 87.78 & 64.76 & 29.47 \\
    \midrule
    \multirow{5}{*}{Tuning-based} & TableLlama-7B & 53.32 & 83.88 & 82.53 & 58.73 & 15.73 \\
          & TableGPT2-7B & 62.46 & 80.43 & 78.87 & 65.42 & 17.69 \\
          & Table-R1 & \underline{74.86} & 96.02 & 87.17 & 69.24 & \underline{41.27} \\
    \cmidrule{2-7}          
          & TableMind & \textbf{76.82} & \textbf{99.27} & \textbf{91.85} & \textbf{71.95} & \textbf{42.02} \\
          & Relative Improvement & (+2.61) & (+1.26) & (+3.28) & (+2.24) & (+1.82) \\
    \bottomrule
    \end{tabular}%
  \label{tab:main result}%
\end{table*}%

%% file: table/2-pure-sft.tex
\begin{table}[t]
  \centering
  \caption{Results from testing different training configurations to find the optimal SFT strategy. }
    \begin{tabular}{ccccc}
    \toprule
    Epoch & Sample Size & Wikitq & TabMWP & TabFact \\
    \midrule
    \multirow{3}[2]{*}{1} & 100   & 52.52 & 62.49 & 70.16 \\
          & 150   & 53.65 & 62.63 & 71.32 \\
          & 200   & 55.40 & 68.42 & 77.15 \\
    \midrule
    \multirow{3}[2]{*}{2} & 100   & 50.12 & 60.62 & 70.12 \\
          & 150   & 50.42 & 61.48 & 70.29 \\
          & 200   & 51.69 & 62.28 & 73.62 \\
    \bottomrule
    
    \end{tabular}%
  \label{tab:pure sft}%
  \vspace{-0.2in}
\end{table}%

%% file: table/3-Hyperparameter.tex
\begin{table}[t]
  \centering
  \caption{Hyperparameter sensitivity analysis for TableMind. This table shows the performance impact of the maximum number of tool-call turns and rollout Temperature.}
    \begin{tabular}{ccccc}
    \toprule
    Hyperparameter & Value & Wikitq & TabMWP & TabFact \\
    \midrule
    \multirow{3}[2]{*}{Max Turns} & 1     & 0.7435 & 0.9723 & 0.8983 \\
          & 3     & 0.7682 & 0.9903 & 0.9233 \\
          & 5     & 0.7565 & 0.9856 & 0.9055 \\
    \midrule
    \multirow{3}[2]{*}{Temperature} & 0.6   & 0.7335 & 0.9658 & 0.8843 \\
          & 0.8   & 0.7551 & 0.9884 & 0.9125 \\
          & 1     & 0.7682 & 0.9903 & 0.9233 \\
    \bottomrule
    \end{tabular}%
  \label{tab: hyperparam}%
  \vspace{-0.2in}
\end{table}%

%% file: 6-Conclusion.tex
\section{Conclusion}
In this paper, we proposed TableMind, an autonomous programmatic agent for table reasoning that moves beyond the prevailing single-step inference paradigm. TableMind enables human-like multi-turn reasoning and precise numerical computation over tables.
TableMind is trained with a two-stage strategy combining supervised fine-tuning and reinforcement learning, allowing the model to autonomously generate and execute programs, verify intermediate results, and iteratively refine its reasoning process. Extensive experiments across diverse benchmarks demonstrate that TableMind consistently outperforms existing baselines, validating the effectiveness of training lightweight agents rather than relying on training-free, workflow-based designs.
More broadly, this work suggests a promising direction for table reasoning by shifting away from rigid single-turn inference or predefined agent pipelines and moving toward learned autonomous agents that natively support structured interaction and reflection. While TableMind focuses on single-table scenarios, extending it to multi-table reasoning, richer tool ecosystems, and improving the sample efficiency of reinforcement fine-tuning remain important directions for future research.

\section*{Acknowledgments}
This research was supported by grants from the National Natural Science Foundation of China (No. 62502486, 62337001), the grants of Provincial Natural Science Foundation of Anhui Province (No.2408085QF193), USTC ResearchFunds of the DoubleFirst-Class Initiative (No. YD2150002501), the Fundamental Research Funds for the Central Universities of China (No. WK2150110032).

%% file: main.bib
@inproceedings{yang2025table,
  title={Table-r1: Inference-time scaling for table reasoning tasks},
  author={Yang, Zheyuan and Chen, Lyuhao and Cohan, Arman and Zhao, Yilun},
  booktitle={Proceedings of the 2025 Conference on Empirical Methods in Natural Language Processing},
  pages={20616--20635},
  year={2025}
}

@article{wang2024survey,
  title={A survey on large language model based autonomous agents},
  author={Wang, Lei and Ma, Chen and Feng, Xueyang and Zhang, Zeyu and Yang, Hao and Zhang, Jingsen and Chen, Zhiyuan and Tang, Jiakai and Chen, Xu and Lin, Yankai and others},
  journal={Frontiers of Computer Science},
  volume={18},
  number={6},
  pages={186345},
  year={2024},
  publisher={Springer}
}

@inproceedings{yao2023react,
  title={React: Synergizing reasoning and acting in language models},
  author={Yao, Shunyu and Zhao, Jeffrey and Yu, Dian and Du, Nan and Shafran, Izhak and Narasimhan, Karthik and Cao, Yuan},
  booktitle={International Conference on Learning Representations (ICLR)},
  year={2023}
}

@article{cheng2025survey,
  title={A survey on table mining with large language models: Challenges, advancements and prospects},
  author={Cheng, Mingyue and Mao, Qingyang and Liu, Qi and Zhou, Yitong and Li, Yupeng and Wang, Jiahao and Lin, Jiaying and Cao, Jiawei and Chen, Enhong},
  journal={Authorea Preprints},
  year={2025},
  publisher={Authorea}
}

@article{hollmann2025accurate,
  title={Accurate predictions on small data with a tabular foundation model},
  author={Hollmann, Noah and M{\"u}ller, Samuel and Purucker, Lennart and Krishnakumar, Arjun and K{\"o}rfer, Max and Hoo, Shi Bin and Schirrmeister, Robin Tibor and Hutter, Frank},
  journal={Nature},
  volume={637},
  number={8045},
  pages={319--326},
  year={2025},
  publisher={Nature Publishing Group UK London}
}

@inproceedings{jin2023tab,
  title={Tab-cot: Zero-shot tabular chain of thought},
  author={Ziqi, Jin and Lu, Wei},
  booktitle={Findings of the Association for Computational Linguistics: ACL 2023},
  pages={10259--10277},
  year={2023}
}

@article{mao2024potable,
  title={PoTable: Towards Systematic Thinking via Stage-oriented Plan-then-Execute Reasoning on Tables},
  author={Mao, Qingyang and Liu, Qi and Li, Zhi and Cheng, Mingyue and Zhang, Zheng and Li, Rui},
  journal={arXiv preprint arXiv:2412.04272},
  year={2024}
}

@article{wang2024chain,
  title={Chain-of-table: Evolving tables in the reasoning chain for table understanding},
  author={Wang, Zilong and Zhang, Hao and Li, Chun-Liang and Eisenschlos, Julian Martin and Perot, Vincent and Wang, Zifeng and Miculicich, Lesly and Fujii, Yasuhisa and Shang, Jingbo and Lee, Chen-Yu and others},
  journal={arXiv preprint arXiv:2401.04398},
  year={2024}
}

@inproceedings{zhang2023tablellama,
  title={Tablellama: Towards open large generalist models for tables},
  author={Zhang, Tianshu and Yue, Xiang and Li, Yifei and Sun, Huan},
  booktitle={Proceedings of the 2024 Conference of the North American Chapter of the Association for Computational Linguistics: Human Language Technologies (Volume 1: Long Papers)},
  pages={6024--6044},
  year={2024}
}

@article{su2024tablegpt2,
  title={Tablegpt2: A large multimodal model with tabular data integration},
  author={Su, Aofeng and Wang, Aowen and Ye, Chao and Zhou, Chen and Zhang, Ga and Chen, Gang and Zhu, Guangcheng and Wang, Haobo and Xu, Haokai and Chen, Hao and others},
  journal={arXiv preprint arXiv:2411.02059},
  year={2024}
}

@inproceedings{herzig2020tapas,
  title={TaPas: Weakly Supervised Table Parsing via Pre-training},
  author={Herzig, Jonathan and Nowak, Pawel Krzysztof and M{\"u}ller, Thomas and Piccinno, Francesco and Eisenschlos, Julian},
  booktitle={Proceedings of the 58th Annual Meeting of the Association for Computational Linguistics},
  year={2020},
  organization={Association for Computational Linguistics}
}

@article{koroteev2021bert,
  title={BERT: a review of applications in natural language processing and understanding},
  author={Koroteev, Mikhail V},
  journal={arXiv preprint arXiv:2103.11943},
  year={2021}
}

@inproceedings{wang2021tuta,
  title={Tuta: Tree-based transformers for generally structured table pre-training},
  author={Wang, Zhiruo and Dong, Haoyu and Jia, Ran and Li, Jia and Fu, Zhiyi and Han, Shi and Zhang, Dongmei},
  booktitle={Proceedings of the 27th ACM SIGKDD Conference on Knowledge Discovery \& Data Mining},
  pages={1780--1790},
  year={2021}
}

@inproceedings{gu2022pasta,
  title={PASTA: table-operations aware fact verification via sentence-table cloze pre-training},
  author={Gu, Zihui and Fan, Ju and Tang, Nan and Nakov, Preslav and Zhao, Xiaoman and Du, Xiaoyong},
  booktitle={Proceedings of the 2022 conference on empirical methods in natural language processing},
  pages={4971--4983},
  year={2022}
}

@article{zhang2024tablellm,
  title={TABLELLM: Enabling Tabular Data Manipulation by LLMs in Real Office Usage Scenarios},
  author={Zhang, Xiaokang and Luo, Sijia and Zhang, Bohan and Ma, Zeyao and Zhang, Jing and Li, Yang and Li, Guanlin and Yao, Zijun and Xu, Kangli and Zhou, Jinchang and others}
}

@article{zha2023tablegpt,
  title={Tablegpt: Towards unifying tables, nature language and commands into one gpt},
  author={Zha, Liangyu and Zhou, Junlin and Li, Liyao and Wang, Rui and Huang, Qingyi and Yang, Saisai and Yuan, Jing and Su, Changbao and Li, Xiang and Su, Aofeng and others},
  journal={arXiv preprint arXiv:2307.08674},
  year={2023}
}

@article{guo2025deepseek,
  title={Deepseek-r1: Incentivizing reasoning capability in llms via reinforcement learning},
  author={Guo, Daya and Yang, Dejian and Zhang, Haowei and Song, Junxiao and Zhang, Ruoyu and Xu, Runxin and Zhu, Qihao and Ma, Shirong and Wang, Peiyi and Bi, Xiao and others},
  journal={arXiv preprint arXiv:2501.12948},
  year={2025}
}

@inproceedings{gou2023critic,
  title={CRITIC: Large Language Models Can Self-Correct with Tool-Interactive Critiquing},
  author={Gou, Zhibin and Shao, Zhihong and Gong, Yeyun and Yang, Yujiu and Duan, Nan and Chen, Weizhu and others},
  booktitle={The Twelfth International Conference on Learning Representations}
}

@inproceedings{pasupat2015compositional,
  title={Compositional semantic parsing on semi-structured tables},
  author={Pasupat, Panupong and Liang, Percy},
  booktitle={Proceedings of the 53rd Annual Meeting of the Association for Computational Linguistics and the 7th International Joint Conference on Natural Language Processing (Volume 1: Long Papers)},
  pages={1470--1480},
  year={2015}
}

@inproceedings{lu2022dynamic,
  title={Dynamic prompt learning via policy gradient for semi-structured mathematical reasoning},
  author={Lu, P and Qiu, L and Chang, KW and Wu, YN and Zhu, SC and Rajpurohit, T and Clark, K and Kalyan, A},
  year={2023},
  organization={International Conference on Learning Representations (ICLR 2023)}
}

@inproceedings{chen2019tabfact,
  title={TabFact: A Large-scale Dataset for Table-based Fact Verification},
  author={Chen, Wenhu and Wang, Hongmin and Chen, Jianshu and Zhang, Yunkai and Wang, Hong and Li, Shiyang and Zhou, Xiyou and Wang, William Yang},
  booktitle={International Conference on Learning Representations}
}

@article{hinton2015distilling,
  title={Distilling the Knowledge in a Neural Network},
  author={Hinton, Geoffrey and Vinyals, Oriol and Dean, Jeff},
  journal={stat},
  volume={1050},
  pages={9},
  year={2015}
}

@article{shao2024deepseekmath,
  title={DeepSeekMath: Pushing the Limits of Mathematical Reasoning in Open Language Models},
  author={Shao, Zhihong and Wang, Peiyi and Zhu, Qihao and Xu, Runxin and Song, Junxiao and Zhang, Mingchuan and Li, YK and Wu, Y and Guo, Daya}
}

@inproceedings{zhang2020summarizing,
  title={Summarizing and exploring tabular data in conversational search},
  author={Zhang, Shuo and Dai, Zhuyun and Balog, Krisztian and Callan, Jamie},
  booktitle={Proceedings of the 43rd International ACM SIGIR Conference on Research and Development in Information Retrieval},
  pages={1537--1540},
  year={2020}
}

@inproceedings{cheng2021hitab,
  title={Hitab: A hierarchical table dataset for question answering and natural language generation},
  author={Cheng, Zhoujun and Dong, Haoyu and Wang, Zhiruo and Jia, Ran and Guo, Jiaqi and Gao, Yan and Han, Shi and Lou, Jian-Guang and Zhang, Dongmei},
  booktitle={Proceedings of the 60th Annual Meeting of the Association for Computational Linguistics (Volume 1: Long Papers)},
  pages={1094--1110},
  year={2022}
}

@inproceedings{zhai2024large,
  title={Large language models and future of information retrieval: opportunities and challenges},
  author={Zhai, ChengXiang},
  booktitle={Proceedings of the 47th international ACM SIGIR conference on research and development in information retrieval},
  pages={481--490},
  year={2024}
}

@article{huang2024understanding,
  title={Understanding the planning of LLM agents: A survey},
  author={Huang, Xu and Liu, Weiwen and Chen, Xiaolong and Wang, Xingmei and Wang, Hao and Lian, Defu and Wang, Yasheng and Tang, Ruiming and Chen, Enhong},
  journal={arXiv preprint arXiv:2402.02716},
  year={2024}
}

@inproceedings{cheng2024towards,
  title={Towards personalized evaluation of large language models with an anonymous crowd-sourcing platform},
  author={Cheng, Mingyue and Zhang, Hao and Yang, Jiqian and Liu, Qi and Li, Li and Huang, Xin and Song, Liwei and Li, Zhi and Huang, Zhenya and Chen, Enhong},
  booktitle={Companion Proceedings of the ACM Web Conference 2024},
  pages={1035--1038},
  year={2024}
}

@article{hu2025open,
  title={Open-reasoner-zero: An open source approach to scaling up reinforcement learning on the base model},
  author={Hu, Jingcheng and Zhang, Yinmin and Han, Qi and Jiang, Daxin and Zhang, Xiangyu and Shum, Heung-Yeung},
  journal={Advances in Neural Information Processing Systems},
  volume={38},
  pages={162239--162262},
  year={2026}
}

@inproceedings{liu2025understanding,
  title={Understanding R1-Zero-Like Training: A Critical Perspective},
  author={Liu, Zichen and Chen, Changyu and Li, Wenjun and Qi, Penghui and Pang, Tianyu and Du, Chao and Lee, Wee Sun and Lin, Min},
  booktitle={Second Conference on Language Modeling}
}

@article{yue2025does,
  title={Does reinforcement learning really incentivize reasoning capacity in llms beyond the base model?},
  author={Chen, Zhiqi and Lu, Rui and Zhao, Andrew and Wang, Zhaokai and Yue, Yang and Song, Shiji and Huang, Gao},
  journal={Advances in Neural Information Processing Systems},
  volume={38},
  pages={57654--57689},
  year={2026}
}

@inproceedings{zhao2024tapera,
  title={TaPERA: Enhancing faithfulness and interpretability in long-form table QA by content planning and execution-based reasoning},
  author={Zhao, Yilun and Chen, Lyuhao and Cohan, Arman and Zhao, Chen},
  booktitle={Proceedings of the 62nd Annual Meeting of the Association for Computational Linguistics (Volume 1: Long Papers)},
  pages={12824--12840},
  year={2024}
}

@inproceedings{ye2023large,
  title={Large language models are versatile decomposers: Decomposing evidence and questions for table-based reasoning},
  author={Ye, Yunhu and Hui, Binyuan and Yang, Min and Li, Binhua and Huang, Fei and Li, Yongbin},
  booktitle={Proceedings of the 46th international ACM SIGIR conference on research and development in information retrieval},
  pages={174--184},
  year={2023}
}

@inproceedings{chen2022convfinqa,
  title={Convfinqa: Exploring the chain of numerical reasoning in conversational finance question answering},
  author={Chen, Zhiyu and Li, Shiyang and Smiley, Charese and Ma, Zhiqiang and Shah, Sameena and Wang, William Yang},
  booktitle={Proceedings of the 2022 conference on empirical methods in natural language processing},
  pages={6279--6292},
  year={2022}
}

@inproceedings{pampari2018emrqa,
  title={emrqa: A large corpus for question answering on electronic medical records},
  author={Pampari, Anusri and Raghavan, Preethi and Liang, Jennifer and Peng, Jian},
  booktitle={Proceedings of the 2018 conference on empirical methods in natural language processing},
  pages={2357--2368},
  year={2018}
}

@inproceedings{wang2024tabletime,
  title={Tabletime: Reformulating time series classification as training-free table understanding with large language models},
  author={Wang, Jiahao and Cheng, Mingyue and Mao, Qingyang and Zhou, Yitong and Wang, Daoyu and Liu, Qi and Xu, Feiyang and Li, Xin},
  booktitle={Proceedings of the 34th ACM International Conference on Information and Knowledge Management},
  pages={3009--3019},
  year={2025}
}

@inproceedings{wang2025can,
  title={Can slow-thinking llms reason over time? empirical studies in time series forecasting},
  author={Cheng, Mingyue and Wang, Jiahao and Wang, Daoyu and Tao, Xiaoyu and Liu, Qi and Chen, Enhong},
  booktitle={Proceedings of the Nineteenth ACM International Conference on Web Search and Data Mining},
  pages={99--110},
  year={2026}
}

@article{masterman2024landscape,
  title={The landscape of emerging ai agent architectures for reasoning, planning, and tool calling: A survey},
  author={Masterman, Tula and Besen, Sandi and Sawtell, Mason and Chao, Alex},
  journal={arXiv preprint arXiv:2404.11584},
  year={2024}
}

@article{zhang2025star,
  title={STaR: Towards Cognitive Table Reasoning via Slow-Thinking Large Language Models},
  author={Zhang, Huajian and Cheng, Mingyue and Luo, Yucong and Tao, Xiaoyu},
  journal={arXiv preprint arXiv:2511.11233},
  year={2025}
}

@inproceedings{chen2021finqa,
  title={Finqa: A dataset of numerical reasoning over financial data},
  author={Chen, Zhiyu and Chen, Wenhu and Smiley, Charese and Shah, Sameena and Borova, Iana and Langdon, Dylan and Moussa, Reema and Beane, Matt and Huang, Ting-Hao and Routledge, Bryan R and others},
  booktitle={Proceedings of the 2021 Conference on Empirical Methods in Natural Language Processing},
  pages={3697--3711},
  year={2021}
}

@article{zhou2025benchmarking,
  title={Benchmarking Multimodal LLMs on Recognition and Understanding over Chemical Tables},
  author={Zhou, Yitong and Cheng, Mingyue and Mao, Qingyang and Luo, Yucong and Liu, Qi and Li, Yupeng and Zhang, Xiaohan and Liu, Deguang and Li, Xin and Chen, Enhong},
  journal={arXiv preprint arXiv:2506.11375},
  year={2025}
}

@article{cheng2025agent,
  title={Agent-R1: Training Powerful LLM Agents with End-to-End Reinforcement Learning},
  author={Cheng, Mingyue and Ouyang, Jie and Yu, Shuo and Yan, Ruiran and Luo, Yucong and Liu, Zirui and Wang, Daoyu and Liu, Qi and Chen, Enhong},
  journal={arXiv preprint arXiv:2511.14460},
  year={2025}
}

@article{tao2025values,
  title={From Values to Tokens: An LLM-Driven Framework for Context-aware Time Series Forecasting via Symbolic Discretization},
  author={Tao, Xiaoyu and Zhang, Shilong and Cheng, Mingyue and Liu, Qi and Wang, Daoyu and Pan, Bokai and Pan, Tingyue and Zhang, Changqing and Wang, Shijin}
}

@inproceedings{hollmanntabpfn,
  title={TabPFN: A Transformer That Solves Small Tabular Classification Problems in a Second},
  author={Hollmann, Noah and M{\"u}ller, Samuel and Eggensperger, Katharina and Hutter, Frank},
  booktitle={The Eleventh International Conference on Learning Representations}
}
